\documentclass{icbo}

\usepackage{url}
\usepackage{tawny}
\usepackage{omn}

\usepackage[pdf]{graphviz}

\lstnewenvironment{code}[1][]%
{\lstset{style=tawnystyle,frame=shadowbox,#1}}{}
\lstMakeShortInline[style=tawnystyle]|

\lstnewenvironment{omn}[1][]%
{\lstset{style=omnstyle,frame=shadowbox,#1}}{}

\begin{document}

\title[Facets, Tiers and Gems]{Facets, Tiers and Gems: Ontology Patterns for
  Hypernormalisation}
\author[Lord and Stevens]{Phillip Lord\,$^1$\footnote{To whom correspondence should be
    addressed: phillip.lord@newcastle.ac.uk} and Robert Stevens$^2$}
\address{$^1$School of Computing Science,\\
  Newcastle University,\\
  Newcastle-upon-Tyne\\
  $^2$School of Computer Science,\\
  University of Manchester,\\
  Manchester
}

\maketitle

\begin{abstract}

  There are many methodologies and techniques for easing the task of
  ontology building. Here we describe the intersection of two of
  these: ontology normalisation and fully programmatic ontology
  development. The first of these describes a standardized
  organisation for an ontology, with singly inherited
  \textit{self-standing} entities, and a number of small taxonomies of
  \textit{refining} entities. The former are described and defined in
  terms of the latter and used to manage the polyhierarchy of the
  self-standing entities. Fully programmatic development is a
  technique where an ontology is developed using a domain-specific
  language within a programming language, meaning that as well
  defining ontological entities, it is possible to add arbitrary
  patterns or new syntax within the same environment. We describe how
  new patterns can be used to enable a new style of ontology
  development that we call \emph{hypernormalisation}.

\end{abstract}

\section{Introduction}

Building ontologies is a difficult and time-consuming business for a number of
reasons: from an abstract point-of-view knowledge about the domain can be
difficult to gather, to understand and to represent ontologically; more,
immediately, ontologies, especially those with a complex representation, can
be taxing to describe and define consistently, to update, expand or change when that
representation needs to change.

There have been numerous attempts to simplify and clarify this process including:
the development of methodologies such as OntoClean that defines a set of
meta-properties that can inform ontological modelling~\citep{Guarino_2002};
upper ontologies such as DOLCE or BFO~\citep{Grenon2004} that provide a
pre-made upper classification. 

Another approach that can leverage both of these techniques is
ontology normalisation~\citep{rector_2002}. Originally intended as a
mechanism for ``untangling'' existing hierarchies or classifications
being reused as the basis for an ontology, it also has significant use
as a pattern for building ontologies \emph{de novo}.

Broadly, a normalised ontology is defined using a skeleton that is a strict
tree (i.e. not a acyclic graph) of concepts differentiated using an
inheritance (i.e. not a partonomy) relationship. These are further split into:
a set of \textit{self-standing entities} in which children are disjoint from
each other, but do not cover the parent, and \textit{partitioning} or
\textit{refining} concepts that form closed, covering and disjoint
hierarchies. Building an ontology in this way, allows the ontology developer
to exploit the reasoner to build a polyhierarchy by using classes that define
the self-standing entity in terms of the refining partitions. Polyhierarchies
are difficult to build manually, as human ontology developers, no matter how
good their domain knowledge, find it hard to ensure all possible parents of an
entity are taken into account. The normalisation approach uses defined classes
and reasoning to remove this chore. Creating the tree of self-standing
entities still, however, remains as a task for the developer. The
normalisation approach can significantly increase the robustness and reduce
the work of manual maintenance~\citep{wroe_2003}. In this latter form,
ontological normalisation has been widely, if implicitly, used.

While the term ``ontology normalisation'' has been borrowed, somewhat
metaphorically, from database engineering, the process of building ontologies
using a set of standard design patterns has a rather more direct relationship
to the software engineering equivalent. By reusing a standard set of patterns,
it is possible to build an ontology both rapidly, and consistently. This has
manifested itself in a number of different ways, with a number of different
tools, such as TermGenie~\citep{Dietze_2014}, or
Populus~\citep{Jupp_Wolstencroft_Stevens_2011} which can generate ontologies
according to a pattern.

We have previously described a fully programmatic methodology for ontology
development~\citep{lord_semantic_2013}, using the Tawny-OWL environment. This
is built around the programming language Clojure and enables the ontology to
take advantage of all the features of a programming language and its
environment, including unit testing~\citep{how_what_why_2015}, build,
evaluation and, of course, pattern-driven development by simple use of
functions~\citep{pattern_driven_approach_2013}.  With respect to patterns, this
environment has several advantages. First, and unlike tools such as Populous
and OPPL~\citep{aranguren_Stevens_Antezana_2009}, patterns are developed in the
same environment and syntax as simple ontology concepts; it is, therefore, as
easy to define a pattern as it is to define a class. Second, being based on
Clojure, a language which is homoiconic and has very little syntax of its own,
it is possible to build arbitrary syntactic constructions to represent
patterns in a way that is both convenient and attractive to the developer.

In this paper, we describe an extension of the normalisation technique that we
call \emph{hypernormalisation}. This technique is typified by the (near or
complete) absence of asserted hierarchy among the self-standing entities. We
describe how this allows construction of an exemplar ontology of
amino-acids~\citep{greycite9379}. We then move on to describe recent
developments in the Tawny-OWL environment, including the definition of two new
design patterns, the tier and the facet, and one syntactic abstraction, the
gem, can be used to enable hypernormalised ontology development. Finally, we
discuss the application of this approach to other ontologies.

\section{Hypernormalisation and Amino Acids}

\begin{figure*}
\digraph[scale=0.5]{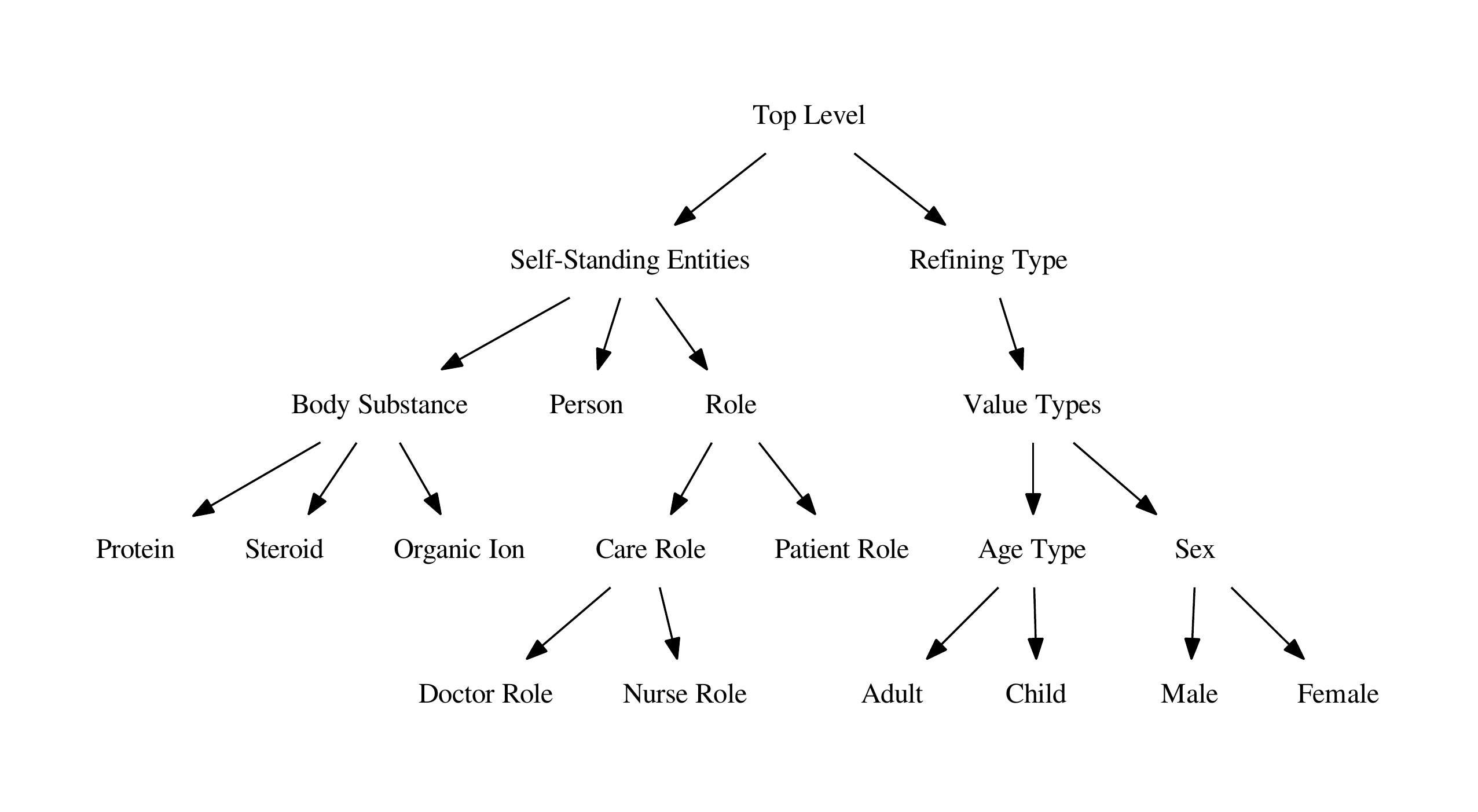}{
  node[shape=plaintext]
    
  {
    tl[label = "Top Level"]
    sse[label="Self-Standing Entities"]
    bs[label="Body Substance"]
  }

  tl-> sse
  sse -> Person
  sse -> bs

  bs -> Protein
  bs -> Steroid
  bs -> "Organic Ion"

  {
    rt[label="Refining Type"]
    at[label="Age Type"]
    cr[label="Care Role"]
    vt[label="Value Types"]
  }
  sse -> Role
  Role -> cr
  Role -> "Patient Role"

  cr -> "Doctor Role"
  cr -> "Nurse Role"

  tl -> rt
  rt -> vt
  vt -> Sex
  Sex -> Male
  Sex -> Female

  vt -> at
  at -> Adult
  at -> Child
}

\label{fig:hyper}
\caption{A normalised ontology slightly modified from \citet{rector_2002}.
  The graph does not necessarily reflect subsumption, see text for details.}
\end{figure*}

Normalisation is a methodology that aims to disentangle an ontological
structure, in the process managing its maintainability, utility and
expressivity of the ontology generated. To achieve this, the ontology is split
into two main hierarchies: self-standing entities and refining types, see
Figure~\ref{fig:hyper} for an example. The self-standing hierarchy contains
entities with a central hierarchy or \emph{skeleton}. In this part of the
ontology, we would expect that hierarchy contains levels that are
not-exhaustive -- that is the children do not cover the parents, and parents
are not closed to new children. This is contrasted by the refining hierarchy
that consists of classes that are exhaustive; in many cases, children
will be non-overlapping and, therefore, disjoint. This is not to say that the
refining types hierarchy are necessarily complete: in Figure~\ref{fig:hyper}, for
example, the representation of \texttt{Sex} is too simple for many medical
uses, but might be sufficient for a customer relations system. In general, the
self-standing entities will be defined in terms of the refining types, while
polyhierarchical relationships between the self-standing entities will be
determined through use of a reasoner.

This form of ontology development is quite different from an upper ontology
and agnostic to the choice of upper ontology or none. While
\citet{rector_2002} suggests only that self-standing entities and refining
types should be ``made clear by some mechanism''; in OWL, it could be an upper
ontological term, or an annotation.

We next introduce the amino-acid, used here as an exemplar, which defines the
biological amino-acids in terms of the physiochemical properties most relevant
to their biological role. It is a structurally interesting ontology because it
is normalised, with a clear and clean separation between the self-standing
entities and the five refining concepts. It is rather more than this, though;
the self-standing entities are split into only three sets: the amino-acids
themselves (e.g. Alanine); a (very large) set of defined classes describing the
refined types of amino-acid (e.g. Small Neutral Amino Acid); and, finally, the
single class Amino Acid. Or, stated alternatively, it contains no skeleton
hierarchy at all, and all relationships between the self-standing classes are
arrived at through reasoning. This is particularly relevant for the amino acid
ontology as it contains over 500 defined classes, with subsumption
relationships to the amino acids and between themselves. Maintaining this
form of ontology by hand would be impractical.

We call this style of ontology development \textit{hypernormalised}. We
believe that it is a natural extension of normalisation. Rector notes, for
example, that the choice of aspect to form the skeleton is ``to some degree
arbitrary'', but that they should be rigid (after
OntoClean~\citep{Guarino_2002}) and pragmatically stable (i.e. unlikely to
change during the evolution of the ontology). Both of these are, however, true for
all the refining concepts in the amino-acids. In short, not only is the
choice of skeleton arbitrary it is actually unnecessary and brings no further
utility to the ontology than that which can be achieved by use of reasoning.

We note that the distinction between normalisation and
hypernormalisation is not absolute, but one of degree; we are simply
describing the tendency toward an ontology with an flat asserted
hierarchy.

Having introduced the notions of a hypernormalised ontology, we next consider
a set of new patterns in Tawny-OWL that enable this style of ontology development.

\begin{figure*}
\digraph[scale=0.5]{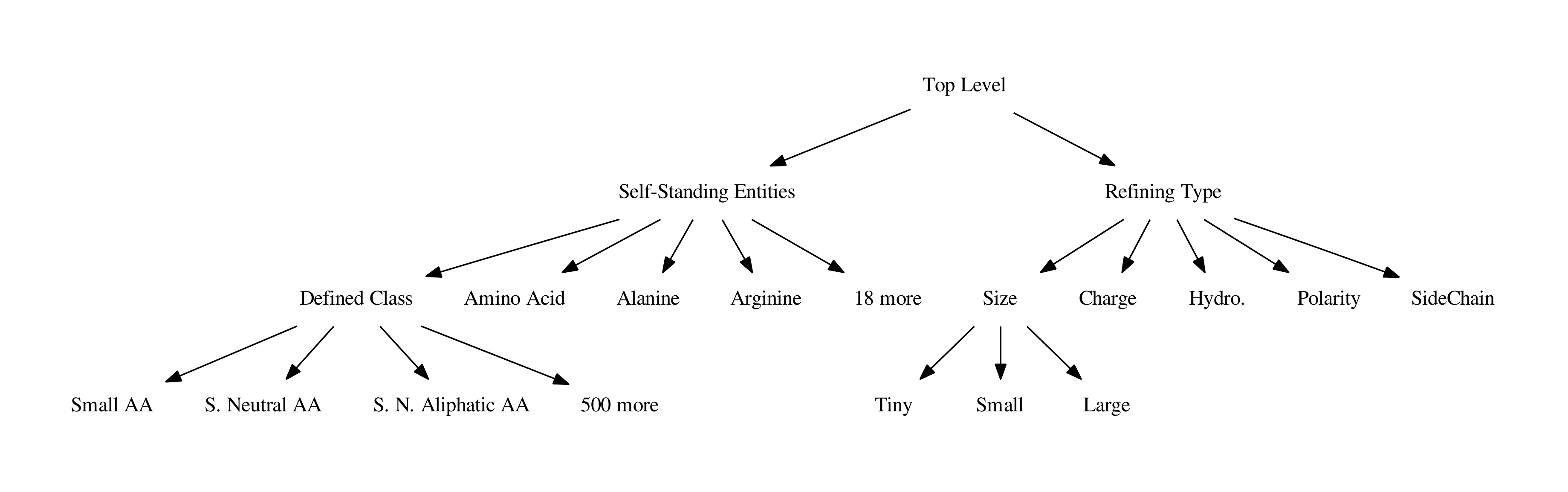}{
  node[shape=plaintext]
    
  {
    tl[label = "Top Level"]
    sse[label="Self-Standing Entities"]
    rt[label="Refining Type"]
    dc[label="Defined Class"]
  }
  tl -> sse

  sse -> "Amino Acid"
  sse -> Alanine
  sse -> Arginine
  sse -> "18 more"

  sse -> dc
  
  dc -> "Small AA"
  dc -> "S. Neutral AA"
  dc -> "S. N. Aliphatic AA"
  dc -> "500 more"
  
  tl -> rt

  rt -> Size
  rt -> Charge
  rt -> "Hydro."
  rt -> Polarity
  rt -> SideChain

  Size -> Tiny
  Size -> Small
  Size -> Large
}
\label{fig:amino}
\caption{A hypernormalised ontology representing the amino-acids using the
  same terminology as Figure~\ref{fig:hyper}. Some labels have been
  abbreviated.}
\end{figure*}

\section{Patternising and Tawny-OWL}

The Tawny-OWL environment~\citep{lord_semantic_2013} and its ability to
support patterns~\citep{pattern_driven_approach_2013} has been described
elsewhere in detail; here, we provide a quick overview, so that the rest of
the paper is clear. Tawny-OWL is implemented as a DSL (domain-specific
language) in Clojure, which is a Lisp-like language implemented in Java, and
running on the Java virtual machine. Tawny-OWL itself wraps the OWL
API~\citep{owlapi_2011}; this is the same library that underpins Protègè, and
from it, Tawny-OWL gains much of its functionality.  Simple sections of the
ontology can be generated using a syntax based on a ``lispified'' version
of Manchester OWL Notation; for example, the following code:

\begin{code}
(defclass A
  :super B)
\end{code}

This declares a new class |A| that has the pre-existing class |B| as a
superclass~\footnote{See \citet{greycite23928} an explanation of why
  \lstinline{:super} is used rather than \lstinline{:subclass}.} which
in Manchester OWL notation would be expressed as:

\begin{omn}
Class: o:A
   SubClassOf: 
        o:B
\end{omn}

This code is entirely valid Clojure and can be evaluated in any
Clojure environment, such as CIDER/Emacs or Cursive/IntelliJ. It is
also possible to define new patterns: for example the following
pattern definition:

\begin{code}
(defn some-only [property & clazzes]
  (list (some property clazzes)
        (only property (or clazzes))))
\end{code}

defines the |some-only| pattern which generates a set of existential
restrictions and one universal with the union of the existential fillers as
its filler, which implements the ontological \emph{closure} pattern.  This is
a function definition in Clojure terms: |defn| introduces the function,
|property & clazzes| is the argument list, |some|, |only| and |or| are
functions provided by Tawny-OWL and |list| returns, prosaically, a
list\footnote{The function shown here is a slightly simplified version of one
  provided in Tawny-OWL.}. Critically, it is possible to define this pattern
in the same environment, or side-by-side in the same file as a simple class
definition; with Tawny-OWL it is as easy to define a class, as to define and
use a new pattern. Ontologies such as the Karyotype ontology make extensive
use of this facility moving freely between ontology and pattern definitions,
as well as literal data structures, utility functions and unit
tests~\citep{2013arXiv1305.3758W}.

Tawny-OWL is now a mature and used software product; the first alpha
release of Tawny-OWL was in Nov 2012, first full release, Nov 2013,
followed by four point releases to 2016. This paper describes mostly
the upcoming v2.0 release, although some of the features described
were available in earlier versions.

\section{The Value Partition}

A common pattern for building a normalised ontology is called the \emph{value
  partition}. This pattern~\citep{rector2005} addresses the problem of the
ontological modelling of a continuous range. For example, in modelling the
amino-acids, we can consider the concept of |Size|; this could be described
directly using the molecular weight of the amino-acid. However, for the
purpose of the amino-acids, it is both easy and general practice to split size
into three categories: tiny, small and large. In Tawny-OWL, this can be
achieved straight-forwardly using the |defpartition| function\footnote{For
  those with knowledge of Lisp, this is actually a macro; the main
  implementation is in the \lstinline|value-partition| function. Tawny-OWL
  provides support for implementing syntactic macros whose function is simply
  to allow the use of bare symbols. For those without knowledge of Lisp, the
  distinction is not important!}.

\begin{code}
(defpartition Size
  [Tiny Small Large]
  :domain AminoAcid
  :super PhysioChemicalProperty)
\end{code}

Axiomatically, this expands into: a class |Size|; three subclasses,
|Tiny|, |Small| and |Large|; and, a property |hasSize|. The property
is functional, has range of |Size| and domain of
|AminoAcid|. Expanded, this would be expressed as
follows\footnote{Tawny-OWL also adds annotations which have been elided}:

\begin{omn}
Class: o:Large
    SubClassOf: 
        o:Size
    
Class: o:Size
    EquivalentTo: 
        o:Large or o:Small or o:Tiny
    
    SubClassOf: 
        o:PhysioChemicalProperty
    
Class: o:Small
    SubClassOf: 
        o:Size
    
Class: o:Tiny
    SubClassOf: 
        o:Size
    
DisjointClasses: 
    o:Large,o:Small,o:Tiny
\end{omn}

The subclasses are disjoint and cover the parent. Following the
terminology from \citet{rector_2002}, the value partition is useful
for defining partitioning or refining concepts.

\section{The Tier}

The value partition is a pattern aimed at a specific purpose -- segmenting a
continuous range. In practice, though, we have found that the axiomatization
of this pattern is more generally useful. For example, considering the
amino-acid ontology, it is natural to model the chemistry of the side-chain as
such:

\begin{code}
(defpartition SideChainStructure
  [Aromatic Aliphatic]
  :domain AminoAcid
  :super PhysicoChemicalProperty)
\end{code}

While this is intuitive, ontologically, |SideChainStructure| is actually of a
very different form from |Size|, as it does not reflect a spectrum. Either the
side-chain contains a benzene ring, making it aromatic, or it does not. This
form of partition was also noted in \citet{rector_2002} which includes the
classes |Male| and |Female| which is not a spectrum, at least in this
simplified representation. We introduce here, therefore, the more general
notion of the \emph{tier}: a small set of concepts in a one-deep
hierarchy. The tier function supports a range of options:

\begin{code}
(deftier Charge
   [Positive Neutral Negative]
   :domain AminoAcid
   :super PhysioChemicalProperty
   :suffix true)
\end{code}

The use of |:suffix true| causes a simple change to the naming of the
entities: |Positive| will become |PositiveCharge| which would be
expanded as follows:

\begin{omn}
Class: o:PositiveCharge
    SubClassOf: 
        o:Charge
\end{omn}

Other names are modified equivalently. By default, this will manifest
both when referring to the class in the Tawny-OWL environment, in the
IRI of the concept when serialized as OWL, and in the value of an
annotation on the concepts\footnote{The duplication between the
  annotation and the IRI fragment is there because IRI schemes such as
  numeric style OBO IDs; annotations have been elided for brevity}. In
addition to naming, it is also possible to optionalise: whether or not
the subclasses are disjoint, covering, whether the property is
functional or whether it is created at all.

The tier is a more general pattern than the value-partition; in fact, in the
current version of Tawny-OWL, the latter is defined in terms of the former.

\section{The Facet}

Both the value partition and tier introduce a new object property named after
the tier, and with a range limited to the classes defined within the tier. The
converse is also true; where we use one of the tier classes, such as
|PositiveCharge| it is most likely that we wish to use it with the |hasCharge|
property defined as part. Taken together, we describe the combination of
classes and a property as a \emph{facet}. Facets are a well known technique,
first proposed in a library classification (the Colon
Classification~\citep{ranganathan_1933}, named after the use of ``:'' as a
separator). They are now common-place as seen with facetted browsers used by
many websites for navigation of complex product catalogues.

Tawny-OWL provides explicit support for facets, allowing the association of a
property and a set of classes, as demonstrated by the following code:

\begin{code}
(as-facet
 hasCharge

 Positive Neutral Negative)
\end{code}

The practical implication of this is that we can now use the |facet| function
to return an existential restriction providing just a class. We can
express this programmatically; for example, we might use the |assert|
function provided by Clojure's unit test framework.

\begin{code}
(assert
 (= (some hasCharge Positive)
    (facet Positive)))
\end{code}

By itself, this ability is only slightly more succinct. However, when used
with multiple facetted classes, the advantages become considerably 
clearer, as can be shown by the following assertion.

\begin{code}
(assert
 (= (list (some hasCharge
                Neutral)
          (some hasHydrophobicity
                Hydrophobic)
          (some hasPolarity
                NonPolar)
          (some hasSideChainStructure
                Aliphatic)
          (some hasSize
                Tiny))
    (facet Neutral Hydrophobic NonPolar
           Aliphatic Tiny)))
\end{code}

In addition to succinctness, this pattern also reduces the risk of errors; a
class such as |Tiny| will always be used with its correct property. Without
the use of facets, the ontology developer must achieve this by
hand. It would also be possible to detect the error using reasoning, although this will
only succeed if appropriate range and disjoint restrictions are in the
ontology. The |defpartition| and |deftier| functions, of course, both add
these range and disjoint restrictions and declare their classes as facets of
their properties.

\section{The Gem}

Finally, we define the \emph{gem} that provides a syntactic abstraction for a
class composed entirely or mainly from facets. Following the terminology from
\citet{rector_2002}, this abstraction would be useful mostly for self-standing
concepts. For example, we could define the amino acid alanine using the
following |defgem| statement.

\begin{code}
(defgem Alanine
  :comment "An amino acid with a single
methyl group as a side-chain."
  :facet Neutral Hydrophobic NonPolar
  Aliphatic Tiny)
\end{code}

The other amino-acids can be likewise defined as a series of gems. In fact, the
amino acids are so regular, all having the same five facets, that we use a further
syntactic abstract specific to the amino-acid ontology -- a form of pattern
that we describe as \emph{localized}~\citep{warrender_thesis_2015}. The gem
represents generalised syntax useful for developing any ontology.

\section{On Annotation}

We have previously discussed the relationship between a design methodology
such as normalisation and the use of an upper ontology. The Tawny-OWL patterns
described here are all orthogonal and agnostic to the choice of an upper
ontology or to none. They do not place their entities in any particular part
of the class hierarchy nor define classes outside of those required for the
domain ontology, although they could be easily extended to do so should the
ontology developer require.

However, we agree with \citet{rector_2002} that the use of patterns should
``made clear'' and be explicit within the ontology. For this reason, all of
the patterns described here also make use of annotations, using annotation
properties defined using its own internal annotation ontology. For example,
all entities generated as a result of a pattern such as |deftier| are
explicitly annotated as such. This means that the use of these patterns is
(informally) explicit in the OWL serialization. Tawny-OWL actually uses these
annotations internally, for example, to enable the |facet| functionality by
providing a relationship between the classes and the appropriate object
property. This is a strictly an implementation detail and could have been
achieved without annotations; however, we believe that it shows the value of
having this knowledge explicit in OWL.

\section{Discussion}

In this paper, we describe how we have used Tawny-OWL to provide higher-level
patterns which can be applied to ontology development. The patterns provide
both functionality and syntactic abstraction over the underlying OWL
implementation. In the process, they enable the easy and accurate construction
of ontologies.

More specifically, we demonstrate two new patterns: the tier and the
facet. The tier is an extension of the existing value partition pattern and
can be used for the generation of many small hierarchies that can be used as
refining properties. The facet borrows from the library sciences notion of a
facetted classification, and is used to associate a set of classes with a
specific set of values. This form of classification is very common in
the web; the majority of web stores, for example, offer facetted
browsing, often with the facets changing for different subsections of
the catalogue.

Taken together, these two patterns enable a new form of ontology development,
hypernormalisation, which is an extreme form of normalisation. In this form of
normalisation, we do away with the creation of a tree of self-standing entities
and instead rely on the reasoner to build all the hierarchy. As well as making
the ontologist's task easier, it makes the characteristic that would have been
used to create the tree of self-standing entities explicit in the form of a
refining characteristic. Here, we have described the application of this
methodology to the exemplar amino-acid ontology. Of course, it is dangerous to
extrapolate to generality from an exemplar, but we have also started to apply
hypernormalisation to ontologies of other, more real, domains including clouds
(in the meterological sense), cell lines and a reworking of the Gene
Ontology. The tier has been made generic; it does not require, for
example, that all refining types are closed (i.e. all possibilities
are known in advance) nor disjoint.

Clearly, not all forms of ontology will naturally be represented in a
hypernormalised form. For example, the Karyotype
ontology~\citep{2013arXiv1305.3758W} is far from this form; here, we
define the self-standing concepts and then use reasoning over a set of
defined classes which effectively operate as
facets~\citep{how_what_why_2015}. However, the popularity of the
facetted browsers shows that is possible to use this form of
classification in many areas. We believe that the introduction of the
concept of hypernormalisation and the implementation of it in
Tawny-OWL could have significant implications for the future
development of ontologies.

\bibliographystyle{natbib}

\bibliography{phil_lord_refs,phil_lord_all,russet}

\begin{thebibliography}{}

\bibitem[Dietze {\em et~al.}(2014)Dietze, Berardini, Foulger, Hill, Lomax,
  OsumiSutherland, Roncaglia, and Mungall]{Dietze_2014}
Dietze, H., Berardini, T.~Z., Foulger, R.~E., Hill, D.~P., Lomax, J.,
  OsumiSutherland, D., Roncaglia, P., and Mungall, C.~J. (2014).
\newblock Termgenie - a web application for pattern-based ontology class
  generation.
\newblock {\em Journal of Biomedical Semantics\/}, {\bf 5}(1), 48.

\bibitem[Egana~Aranguren {\em et~al.}(2009)Egana~Aranguren, Stevens, and
  Antezana]{aranguren_Stevens_Antezana_2009}
Egana~Aranguren, M., Stevens, R., and Antezana, E. (2009).
\newblock Transforming the axiomisation of ontologies: The ontology
  pre-processor language.
\newblock {\em Nature Precedings\/}.

\bibitem[Grenon {\em et~al.}(2004)Grenon, Smith, and Goldberg]{Grenon2004}
Grenon, P., Smith, B., and Goldberg, L. (2004).
\newblock {Biodynamic ontology: applying BFO in the biomedical domain.}
\newblock {\em Stud Health Technol Inform\/}, {\bf 102}, 20--38.

\bibitem[Guarino and Welty(2002)Guarino and Welty]{Guarino_2002}
Guarino, N. and Welty, C. (2002).
\newblock Evaluating ontological decisions with ontoclean.
\newblock {\em Commun. ACM\/}, {\bf 45}(2), 61--65.

\bibitem[Horridge and Bechhofer(2011)Horridge and Bechhofer]{owlapi_2011}
Horridge, M. and Bechhofer, S. (2011).
\newblock {The OWL API: A Java API for OWL Ontologies}.
\newblock {\em Semantic Web Journal\/}, {\bf 2}.

\bibitem[Jupp {\em et~al.}(2011)Jupp, Horridge, Iannone, Klein, Owen,
  Schanstra, Wolstencroft, and Stevens]{Jupp_Wolstencroft_Stevens_2011}
Jupp, S., Horridge, M., Iannone, L., Klein, J., Owen, S., Schanstra, J.,
  Wolstencroft, K., and Stevens, R. (2011).
\newblock Populous: a tool for building owl ontologies from templates.
\newblock {\em BMC Bioinformatics\/}, {\bf 13}(Suppl 1), S5.

\bibitem[Lord(2013)Lord]{lord_semantic_2013}
Lord, P. (2013).
\newblock {The Semantic Web takes Wing: Programming Ontologies with Tawny-OWL}.
\newblock {\em {OWLED 2013}\/}.

\bibitem[Lord(2014)Lord]{greycite23928}
Lord, P. (2014).
\newblock Manchester syntax is a bit backward.
\newblock \url{http://www.russet.org.uk/blog/2985}.

\bibitem[Ranganathan(1933)Ranganathan]{ranganathan_1933}
Ranganathan, S. (1933).
\newblock {\em Colon Classification\/}.

\bibitem[Rector(2005)Rector]{rector2005}
Rector, A. (2005).
\newblock Representing specified values in owl: ``value partitions'' and
  ``value sets''.
\newblock W3C Working Group Note.

\bibitem[Rector(2002)Rector]{rector_2002}
Rector, A.~L. (2002).
\newblock Normalisation of ontology implementations: Towards modularity,
  re-use, and maintainability.
\newblock {\em Proceedings Workshop on Ontologies for Multiagent Systems (OMAS)
  in conjunction with European Knowledge Acquisition Workshops\/}.
\newblock Siguenza, Spain.

\bibitem[Stevens and Lord(2012)Stevens and Lord]{greycite9379}
Stevens, R. and Lord, P. (2012).
\newblock Semantic publishing of knowledge about amino acids.
\newblock \url{http://ceur-ws.org/Vol-903/paper-06.pdf}.

\bibitem[Warrender(2015)Warrender]{warrender_thesis_2015}
Warrender, J. (2015).
\newblock {\em The Consistent Representation of Scientific Knowledge:
  Investigations into the Ontology of Karyotypes and Mitochondria.}
\newblock Ph.D. thesis, School of Computing Science, Newcastle University.

\bibitem[Warrender and Lord(2013)Warrender and
  Lord]{pattern_driven_approach_2013}
Warrender, J. and Lord, P. (2013).
\newblock A pattern-driven approach to biomedical ontology engineering.
\newblock {\em SWAT4LS 2013\/}.

\bibitem[{Warrender} and {Lord}(2013){Warrender} and
  {Lord}]{2013arXiv1305.3758W}
{Warrender}, J.~D. and {Lord}, P. (2013).
\newblock {The Karyotype Ontology: a computational representation for human
  cytogenetic patterns}.
\newblock {\em Bio-Ontologies 2013\/}.

\bibitem[{Warrender} and {Lord}(2015){Warrender} and {Lord}]{how_what_why_2015}
{Warrender}, J.~D. and {Lord}, P. (2015).
\newblock {How, What and Why to test an ontology}.

\bibitem[Wroe {\em et~al.}(2003)Wroe, Stevens, Goble, and Ashburner]{wroe_2003}
Wroe, C., Stevens, R., Goble, C., and Ashburner, M. (2003).
\newblock A methodology to migrate the gene ontology to a description logic
  environment using daml+oil.
\newblock Pacific Symposium on Biocomputing.

\end{thebibliography}

\end{document}